\documentclass{hld2023} 
\usepackage{graphicx}

\usepackage{amsmath,amsfonts,bm, physics}

\def\eqref#1{Eq.~(\ref{#1})}

\def\1{\bm{1}}

\def\ru{{\textnormal{u}}}
\def\rv{{\textnormal{v}}}

\def\vh{{\bm{h}}}

\def\vw{{\bm{w}}}
\def\vx{{\bm{x}}}

\def\vz{{\bm{z}}}

\def\mA{{\bm{A}}}
\def\mB{{\bm{B}}}

\def\mG{{\bm{G}}}

\def\mI{{\bm{I}}}

\def\mK{{\bm{K}}}

\def\mW{{\bm{W}}}
\def\mX{{\bm{X}}}
\def\mY{{\bm{Y}}}

\DeclareMathAlphabet{\mathsfit}{\encodingdefault}{\sfdefault}{m}{sl}
\SetMathAlphabet{\mathsfit}{bold}{\encodingdefault}{\sfdefault}{bx}{n}

\def\gN{{\mathcal{N}}}

\def\sS{{\mathbb{S}}}

\def\*#1{\mathbf{#1}}
\def\$#1{\mathcal{#1}}
\def\^#1{\mathbb{#1}}

\newcommand{\E}{\mathbb{E}}

\newcommand{\R}{\mathbb{R}}

\usepackage{mathtools}
\usepackage{amsmath,amsfonts}

\newcommand{\bsigma}{{\boldsymbol{\Sigma}}}

\newcommand{\btheta}{{\boldsymbol{\Theta}}}

\newtheorem{lem}{Lemma} 
\newtheorem{thm}{Theorem}
\newtheorem{cor}{Corollary}
\newtheorem{assum}{Assumption}

\newtheorem{rem}{Remark}

\title[On the Equivalence between Implicit and Explicit NNs]{On the Equivalence between Implicit and Explicit Neural Networks: \\A High-dimensional Viewpoint}

\hldauthor{%
\Name{Zenan Ling} \Email{lingzenan@hust.edu.cn}\\
\Name{Zhenyu Liao} \Email{zhenyu\_liao@hust.edu.cn}\\
\Name{Robert C.\@ Qiu} \Email{caiming@hust.edu.cn}\\
\addr{ Huazhong University of Science and Technology, Wuhan, China } 
}

\begin{document}

\maketitle

\begin{abstract}
Implicit neural networks have demonstrated remarkable success in various tasks. However, there is a lack of theoretical analysis of the connections and  differences between implicit  and explicit networks. In this paper, we study  high-dimensional implicit neural networks and provide the high dimensional  equivalents for the corresponding conjugate kernels and neural tangent kernels. Built upon this, we establish the equivalence between implicit and explicit networks in high dimensions.
\end{abstract}


\section{Introduction}
Implicit neural networks (NNs)~\cite{NEURIPS2019_01386bd6} have recently emerged as a new paradigm in neural network design.
 An implicit NN is equivalent to  an infinite-depth weight-shared explicit NN with input-injection. Unlike explicit NNs,   implicit NNs generate features by  directly solving for the fixed point, rather than
through  layer-by-layer forward propagation.  Moreover, implicit NNs  have the remarkable advantage that gradients  can be computed analytically  only through the fixed point with \emph{implicit differentiation}. Therefore, training implicit NNs only requires constant memory.

Despite the empirical success achieved by implicit NNs~\cite{bai2020multiscale,xie2022optimization}, our theoretical understanding of these models is still limited.  In particular, there is a lack of theoretical analysis of the training dynamics and generalization performance of implicit NNs, and possibly more importantly, whether  these properties can be connected to those of explicit NNs.~\cite{NEURIPS2019_01386bd6} demonstrates that any deep NN can be reformulated as a special  implicit NN. However, it remains unknown whether general implicit NNs have advantages over explicit NNs.   ~\cite{feng2020neural} extends previous neural tangent kernel (NTK) studies to implicit NNs and give the exact expression of the NTK of the ReLU implicit NNs. However, the differences between implicit and explicit NTKs are not analyzed.
Moreover, previous works~\cite{ling2023global,truong2023global} have proved  the global convergence of gradient descent for  training implicit NNs. However,  it is still unclear what distinguishes the training dynamic of implicit NNs and that of explicit NNs.

In this paper, we investigate implicit NNs from a  high-dimensional view. Specifically, we perform a fine-grained asymptotic analysis on the eigenspectra of conjugate kernel (CKs) and NTKs of implicit NNs, which play a fundamental role in the convergence and  generalization high dimensional NNs~\cite{jacot2018neural}.  By considering input data uniformly drawn from the unit sphere, we derive, with recent advances in random matrix theory, high-dimensional (spectral) equivalents for the  CKs and NTKs of implicit NNs, and establish the equivalence between implicit and explicit NNs by matching the coefficients of the corresponding asymptotic spectral equivalents.
Surprisingly, our results reveal that a \textit{single-layer} explicit NN with  carefully designed activations has the same CK or NTK eigenspectra as a ReLU implicit NN, whose depth is essentially \emph{infinite}.


\section{Preliminaries}
\subsection{Implicit  and Explicit NNs}
\paragraph{Implicit NNs.} In this paper, we study a typical implicit neural network, the deep equilibrium model (DEQ)~\cite{NEURIPS2019_01386bd6}. Let $\mX=[\vx_1,\cdots,\vx_n]\in \mathbb{R}^{d \times n}$ denote the  input data. We define a vanilla DEQ with the transform at the $l$-th layer as
	\begin{equation}
	\vh_i^{(l)} = \sqrt{\frac{\sigma_a^2}{m}}\mA \vz_i^{(l-1)}+ \sqrt{\frac{\sigma_b^2}{m}}\mB\vx_i, \quad   \vz_i^{(l)} = \phi(\vh_i^{(l)})
	\label{eq:deql}
	\end{equation}
	where   $\mA \in \mathbb{R}^{m \times m}$  and  $\mB \in \mathbb{R}^{m \times d}$ are weight matrices, $\sigma_a, \sigma_b\in \R$ are  constants, $\phi$ is an element-wise activation, $\vh_i^{(l)}$ is the pre-activation   and $\vz_i^{(l)} \in \mathbb{R}^{m }$ is the output feature of the $l$-th hidden layer corresponding to the input data $\vx_i$. The output  of the last hidden layer is defined by $ \vz_i^* \triangleq \lim_{l \rightarrow \infty} \vz_i^{(l)}$ and we denote the corresponding pre-activation by $\vh_i^*$. Note that $\vz_i^*$ can be calculated by directly solving for the equilibrium point of the following equation 
	\begin{equation}
	   \vz_i^{*} = \phi\left(\sqrt{\frac{\sigma_a^2}{m}}\mA \vz_i^{*}+\sqrt{\frac{\sigma_b^2}{m}}\mB \vx_i \right). 
	   \label{eq:deq}
	\end{equation}
  We are interested in the  conjugate kernel and neural tangent kernel (Implicit-CK and Implicit-NTK, for short) of implicit neural networks defined in~\eqref{eq:deq}. Following~\cite{feng2020neural},
we denote the corresponding Implicit-CK by $\mG^* = \lim_{l\rightarrow\infty}\mG^{(l)}$ where the $(i,j)$-th entry of $\mG^{(l)}$ is defined recursively as
\begin{equation}
    \begin{split}
        &\mG^{(0)}_{ij} = \vx_i^\top\vx_j, \quad
     \boldsymbol{\Lambda}_{ij}^{(l)} = \left[\begin{array}{cc}
          \mG^{(l-1)}_{ii} & \mG^{(l-1)}_{ij} 
          \\
         \mG^{(l-1)}_{ji} & \mG^{(l-1)}_{jj}
     \end{array}\right],\\
     &\mG^{(l)}_{ij} = \sigma_a^2\E_{(\ru,\rv)\sim \gN(0,\boldsymbol{\Lambda}_{ij}^{(l)})}[\phi(\ru)\phi(\rv)] + \sigma_b^2\vx_i^\top\vx_j, \\
     &\dot\mG^{(l)}_{ij} = \sigma_a^2\E_{(\ru,\rv)\sim \gN(0,\boldsymbol{\Lambda}_{ij}^{(l)})}[\phi'(\ru)\phi'(\rv)].
    \end{split}
\end{equation}
And the Implicit-NTK  is defined as $\mK^*=\lim_{l\rightarrow\infty}\mK^{(l)}$ whose the $(i,j)$-th entry  is defined as
 \begin{equation}
     \mK^{(l)}_{ij} = \sum_{h=1}^{l+1}\left(\mG^{(h-1)}_{ij}\prod_{h'=h}^{l+1}\dot\mG^{(h')}_{ij}\right).
 \end{equation}

\paragraph{Explicit Neural Networks.}
We consider a single-layer fully-connected NN model defined as $\mY = \sqrt{\frac{1}{p}}\sigma(\mW\mX)$ where $\mW \in \R^{p\times d}$ is the weight matrix and $\sigma$ is an element-wise activation function.
Let $\vw\sim\gN(0,\mI_d)$, the corresponding Explicit-CK matrix $\bsigma$ and Explicit-NTK matrix $\btheta$ are defined as follows:
 \begin{equation}
     \bsigma = \E_{\vw}[\sigma(\vw^\top\mX)^\top\sigma(\vw^\top\mX)], \quad \btheta= \bsigma+ \left(\mX^\top\mX\right) \odot \E_{\vw}[\sigma'(\vw^\top\mX)^\top\sigma'(\vw^\top\mX)].
     \label{eq:exkernel}
 \end{equation}

\subsection{CKs and NTKs of ReLU Implicit NNs}
We make the following assumptions on the
random initialization, the input data, and activations.
\begin{assum} (\romannumeral1) As $n\rightarrow \infty$, $d/n\rightarrow c \in (0,\infty)$. All data points $\vx_i$, $i\in [n]$, are independent and uniformly sampled from $\sS^{d-1}$. (\romannumeral2) 
$\mA$, $\mB$, and $\mW$  are independent and have i.i.d entries of zero mean, unit variance, and finite fourth kurtosis. Moreover, we require  $\sigma_a^2+\sigma_b^2=1$. (\romannumeral3) The activation $\phi$ of the implicit NN is the normalized ReLU, i.e., $\phi(x) = \sqrt{2}\max(x,0)$. The activation $\sigma$ of the explicit NN is a $C^3$ function.
\label{assum}
\end{assum}
\begin{rem} 
    (\romannumeral1) Despite derived here for uniform distribution on the unit sphere, we conjecture that our results extend the result to more general distributions by using the technique developed in~\cite{fan2020spectra,du2022lossless}. (\romannumeral2) The additional requirement on the variance is to ensure the existence and uniqueness of the fixed point of the NTK and to keep the diagonal  entries of the CK matrix at $1$, see examples in~\cite{feng2020neural}. (\romannumeral3) It is possible to extend our results to implicit NNs with general activations by using the technique proposed in~\cite{truong2023global}. We defer the extension to more general data distributions and activation functions to future work.
\end{rem}
Under Assumptions~\ref{assum}, the limits of Implicit-CK and Implicit-NTK exist, and one can have  precise expressions of $\mG^*$ and $\mK^*$ as follows~\cite{feng2020neural,ling2023global}.
\begin{lem} Let $f(x) =\frac{\sqrt{1-x^2}+\left(\pi-\arccos x\right)x}{\pi}$.
Under Assumptions~\ref{assum}, the fixed point of Implicit-CK $\mG^*_{ij}$ is the root of
\begin{equation}
    \mG^*_{ij}= \sigma_a^2f(\mG^*_{ij})+(1-\sigma_a^2)\vx_i^\top\vx_j.
    \label{eq:imck}
\end{equation}
The limit of Implicit-NTK is 
\begin{equation}
        \mK^*_{ij} = h(\mG^*_{ij}) \triangleq \frac{\mG^*_{ij}}{1-\dot\mG^*_{ij}} \quad \text{where} \quad \dot\mG^*_{ij}\triangleq \sigma_a^2\pi^{-1}(\pi-\arccos(\mG^*_{ij})).
        \label{eq:imntk}
    \end{equation}
\end{lem}

\section{Main Results}
In this section, we prove the high-dimensional  equivalents for CKs and NTKs of implicit and explicit NNs. As a result, by matching the coefficients of the asymptotic spectral equivalents, we establish the equivalence between  implicit and explicit NNs in high dimensions. 
\subsection{Asymptotic Approximations}
\paragraph{CKs.} We begin by defining several quantities that are crucial to our results. Note that the unique fixed point of ~\eqref{eq:imck} exists as long as $\sigma_a^2<1$.
We define the implicit map induced from~\eqref{eq:imck} as $\mG^*_{ij} \triangleq g(\vx_i^\top\vx_j)$.  
Let $\angle^* = g(0)$ be the solution of $\angle^* = \sigma_a^2f(\angle^* )$ when $\vx_i^\top\vx_j=0$. 
Using implicit differentiation, one can obtain that \[g'(0) =\frac{1-\sigma_a^2}{1-\sigma_a^2f'(\angle^*)},\quad
    g''(0) =\frac{\sigma_a^2(1-\sigma_a^2)^2f''(\angle^*)}{(1-\sigma_a^2f'(\angle^*))^3}.\]
    Now we are ready to present the asymptotic equivalent of the Implicit-CK matrix.
\begin{thm}[Asymptotic approximation of Implicit-CKs] Let Assumptions~\ref{assum} hold. As $n, d \rightarrow\infty$,
the Implicit-CK matrix $\mG^*$ defined in~\eqref{eq:imck} can be approximated consistently in operator norm,  by the matrix $\overline{\mG}$, that is $\|\mG^*-\overline{\mG}\|_2\rightarrow0$, where
\begin{equation*}
    \overline{\mG} = \alpha\bm{1}\bm{1}^\top+\beta\mX^\top\mX+ \mu\mI_n,
\end{equation*}
with $\alpha = g(0) + \frac{g''(0)}{2d}$, $ \beta= g'(0)$, and $\mu = g(1)-g(0)-g'(0)$.
\label{thm:ick}
\end{thm}

\begin{thm}[Asymptotic approximation for Explicit-CKs]Let Assumptions~\ref{assum} hold. As $n, d \rightarrow\infty$,
the Explicit-CK matrix $\bsigma$ defined in~\eqref{eq:exkernel} can be approximated consistently in operator norm, by the matrix $\overline{\bsigma}$,  that is $\|\bsigma-\overline{\bsigma}\|_2\rightarrow0$, where
\begin{equation*}
    \overline{\bsigma} = \alpha_1\bm{1}\bm{1}^\top+\beta_1\mX^\top\mX+ \mu_1\mI_n,
\end{equation*}
with $\alpha_1 = \E[\sigma(z)]^2+\frac{\E[\sigma''(z)]^2}{2d} $, $\beta_1=\E[\sigma'(z)]^2$, and $\mu_1 = \E[\sigma^2(z)] -\E[\sigma(z)]^2-\E[\sigma'(z)]^2$, for $z\sim\gN(0,1)$.
\label{thm:eck}
\end{thm}

\paragraph{NTKs.}
For the Implicit-NTK, we define $\mK^*_{ij} = k(\vx_i^\top\vx_j)$, i.e., $k(\vx_i^\top\vx_j) = h(g(\vx_i^\top\vx_j))$, for $i,j\in[n]$. It is easy to check that $k(0) = h(\angle^*)$ and $k(1) = h(g(1))$. Using  implicit differentiation again, we have \[
  k'(0)=  \frac{(1-\sigma_a^2)h'(\angle^*)}{\sigma_a^2f'(\angle^*)-1},\, k''(0)= \frac{(1-\sigma_a^2)^2(h''(\angle^*)-\sigma_a^2f'(\angle^*)h''(\angle^*)+\sigma_a^2h'(\angle^*)f''(\angle^*))}{(1-\sigma_a^2f'(\angle^*))^3}.\] 
  Now we are ready to present the asymptotic equivalent of the Implicit-NTK matrix.
\begin{thm}[Asymptotic approximation for Implicit-NTKs] Let Assumptions~\ref{assum} hold. As $n, d \rightarrow\infty$,
the Implicit-NTK matrix $\mK^*$ defined~\eqref{eq:imntk} in can be approximated consistently in operator norm,  by the matrix $\overline{\mK}$, that is $\|\mK^*-\overline{\mK}\|_2\rightarrow0$, where
\begin{equation*}
    \overline{\mK} = \dot\alpha\bm{1}\bm{1}^\top+\dot\beta\mX^\top\mX+ \dot\mu\mI_n,
\end{equation*}
with $ \dot\alpha = k(0) + \frac{k''(0)}{2d}$, $ \dot\beta= k'(0) $, and $ \dot\mu = k(1)-k(0)-k'(0)$.
\label{thm:intk}
\end{thm}

\begin{thm}[Asymptotic approximation for Explicit-NTKs]Let Assumptions~\ref{assum} hold. As $n, d \rightarrow\infty$,
the Explicit-NTK matrix $\btheta$ defined in~\eqref{eq:exkernel} can be approximated consistently in operator norm,  by the matrix $\overline{\btheta}$,  that is $\|\btheta^*-\overline{\btheta}\|_2\rightarrow0$, where
\begin{equation*}
    \overline{\btheta}= \dot\alpha_1\bm{1}\bm{1}^\top+\dot\beta_1\mX^\top\mX+ \dot\mu_1\mI_n,
\end{equation*}
with $ \dot\alpha_1 = \E[\sigma(z)]^2+\frac{3\E[\sigma''(z)]^2}{2d}$, $\dot\beta_1=2\E[\sigma'(z)]^2$, and $ \dot\mu_1 = \E[\sigma^2(z)] +\E[\sigma'(z)^2]-\E[\sigma(z)]^2-2\E[\sigma'(z)]^2$ for $z\sim\gN(0,1)$.
\label{thm:entk}
\end{thm}
\begin{rem}
    (\romannumeral1) Due to the homogeneity of the ReLU function, the Implicit-CK and the Implicit-NTK are essentially inner product kernel random matrices.  Consequently, Theorem~\ref{thm:ick} and~\ref{thm:intk} can be built upon the results in~\cite{el2010spectrum}. We postpone the study on general activations to future work.
    (\romannumeral2)  The results in Theorem~\ref{thm:eck} and~\ref{thm:entk} generalize those of~\cite{du2022lossless,ali2021random} to the cases of ``non-centred'' activations, i.e., we do not require $\E[\sigma(z)]=0$ for $z\sim\gN(0,1)$.  
\end{rem}
\subsection{The Equivalence between Implicit and Explicit NNs}
\begin{figure}
    \centering
\includegraphics[width=0.95\textwidth]{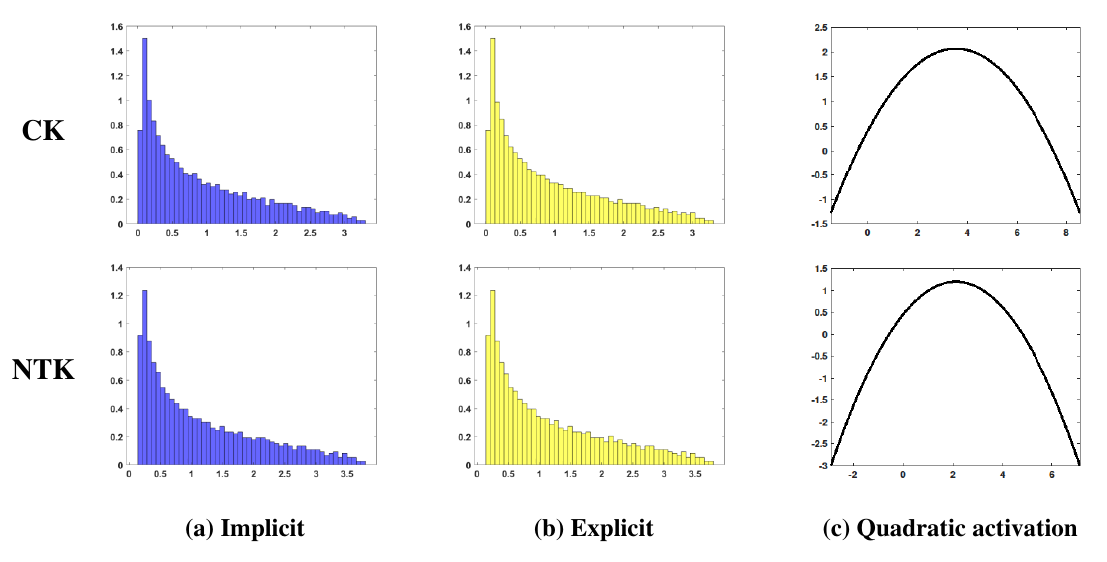}
    \caption{We independently generate $n=1\,000$ data points from the $d=1\,200$-dimensional unit sphere. 
    We use Gaussian initialization and $\sigma_a^2$ is set as $0.2$. Upper: the CK results. Bottom: the NTK results. (a) spectral densities of implicit kernels,  (b) spectral densities of explicit kernels, (c) quadratic activations.}
    \label{fig:ckntk}
\end{figure}
In the following corollary, we show a concrete case of a single-layer explicit NN with an quadratic activation, that matches the  CK or NTK eigenspectra of a ReLU implicit NN. The idea is to utilize the results of 
Theorems~\ref{thm:ick}-\ref{thm:entk} to match the coefficients of the asymptotic equivalents such that $\alpha_1=\alpha, \beta_1=\beta, \mu_1=\mu$, or $\dot\alpha_1=\dot\alpha, \dot\beta_1=\beta, \dot\mu_1=\mu$. We implement  numerical simulations to verify our theory. The numerical results are shown in Figure~\ref{fig:ckntk}.
\begin{cor} 
    We consider a quadratic polynomial activation $\sigma(t) = a_2 t^2 + a_1 t+a_0$. Let Assumptions~\ref{assum} hold. As $n,d\rightarrow\infty$, the Implicit-CK matrix $\mG^*$ defined in~\eqref{eq:imck} can be approximated consistently in operator norm, by the Explicit-CK matrix $\bsigma$ defined in~\eqref{eq:exkernel}, i.e., $\|\mG^*-\bsigma\|_2\rightarrow0$, as long as 
  \[a_2 =\pm\sqrt{\frac{\mu}{2}}\quad a_1 =  \pm\sqrt{\beta}, \quad a_0  = \pm\sqrt{\alpha-\frac{\mu}{d}}-a_2,\]
and the Implicit-NTK matrix $\mK^*$ defined in~\eqref{eq:imntk} can be approximated consistently in operator norm, by the Explicit-NTK matrix $\btheta$ defined in~\eqref{eq:exkernel}, i.e., $\|\mK^*-\btheta\|_2\rightarrow0$, as long as \[a_2 =\pm\sqrt{\frac{\dot\mu}{6}},\quad a_1 =  \pm\sqrt{\frac{\dot\beta}{2}}, \quad a_0  = \pm\sqrt{\dot\alpha-\frac{\dot\mu}{d}}- a_2.\] 
    \label{cor}
\end{cor}
\section{Conclusion}
In this paper, we study the CKs and NTKs of high-dimensional ReLU implicit NNs. We prove the asymptotic spectral equivalents for Implicit-CKs and Implicit-NTKs. Moreover, we establish the equivalence between implicit and explicit NNs by matching the coefficients of the asymptotic spectral equivalents. In particular, we show that  a single-layer explicit NN with carefully designed activations has the same CK or NTK eigenspectra as a ReLU implicit NN. For future work, it would be interesting to extend our analysis to more general data distributions and activation functions.

\paragraph{Acknowledgements} Z.~Liao would like to acknowledge the National Natural Science Foundation of China (via fund NSFC-62206101) and the Fundamental Research Funds for the Central Universities of China (2021XXJS110)  for providing partial support.
R.~C.~Qiu and Z.~Liao  would like to  acknowledge the National Natural Science Foundation of China (via fund NSFC-12141107), the Key Research and Development Program 
of  Hubei (2021BAA037) and of Guangxi (GuiKe-AB21196034).

\bibliography{sample}

\begin{thebibliography}{11}
\providecommand{\natexlab}[1]{#1}
\providecommand{\url}[1]{\texttt{#1}}
\expandafter\ifx\csname urlstyle\endcsname\relax
  \providecommand{\doi}[1]{doi: #1}\else
  \providecommand{\doi}{doi: \begingroup \urlstyle{rm}\Url}\fi

\bibitem[Ali et~al.(2022)Ali, Liao, and Couillet]{ali2021random}
Hafiz~Tiomoko Ali, Zhenyu Liao, and Romain Couillet.
\newblock Random matrices in service of ml footprint: ternary random features
  with no performance loss.
\newblock \emph{ICLR}, 2022.

\bibitem[Bai et~al.(2019)Bai, Kolter, and Koltun]{NEURIPS2019_01386bd6}
Shaojie Bai, J.~Zico Kolter, and Vladlen Koltun.
\newblock Deep equilibrium models.
\newblock In \emph{Advances in Neural Information Processing Systems},
  volume~32. Curran Associates, Inc., 2019.

\bibitem[Bai et~al.(2020)Bai, Koltun, and Kolter]{bai2020multiscale}
Shaojie Bai, Vladlen Koltun, and J~Zico Kolter.
\newblock Multiscale deep equilibrium models.
\newblock \emph{Advances in Neural Information Processing Systems}, 2020.

\bibitem[El~Karoui(2010)]{el2010spectrum}
Noureddine El~Karoui.
\newblock The spectrum of kernel random matrices.
\newblock \emph{The Annuals of Statistics}, 2010.

\bibitem[Fan and Wang(2020)]{fan2020spectra}
Zhou Fan and Zhichao Wang.
\newblock Spectra of the conjugate kernel and neural tangent kernel for
  linear-width neural networks.
\newblock \emph{Advances in neural information processing systems},
  33:\penalty0 7710--7721, 2020.

\bibitem[Feng and Kolter(2020)]{feng2020neural}
Zhili Feng and J~Zico Kolter.
\newblock On the neural tangent kernel of equilibrium models.
\newblock \emph{arxiv}, 2020.

\bibitem[Gu et~al.(2022)Gu, Du, Yuan, Xie, Pu, Qiu, and Liao]{du2022lossless}
Lingyu Gu, Yongqi Du, Zhang Yuan, Di~Xie, Shiliang Pu, Robert Qiu, and Zhenyu
  Liao.
\newblock " lossless" compression of deep neural networks: A high-dimensional
  neural tangent kernel approach.
\newblock \emph{Advances in Neural Information Processing Systems},
  35:\penalty0 3774--3787, 2022.

\bibitem[Jacot et~al.(2018)Jacot, Gabriel, and Hongler]{jacot2018neural}
Arthur Jacot, Franck Gabriel, and Cl{\'e}ment Hongler.
\newblock Neural tangent kernel: Convergence and generalization in neural
  networks.
\newblock \emph{Advances in neural information processing systems}, 31, 2018.

\bibitem[Ling et~al.(2023)Ling, Xie, Wang, Zhang, and Lin]{ling2023global}
Zenan Ling, Xingyu Xie, Qiuhao Wang, Zongpeng Zhang, and Zhouchen Lin.
\newblock Global convergence of over-parameterized deep equilibrium models.
\newblock In \emph{International Conference on Artificial Intelligence and
  Statistics}, pages 767--787. PMLR, 2023.

\bibitem[Truong(2023)]{truong2023global}
Lan~V Truong.
\newblock Global convergence rate of deep equilibrium models with general
  activations.
\newblock \emph{arXiv preprint arXiv:2302.05797}, 2023.

\bibitem[Xie et~al.(2022)Xie, Wang, Ling, Li, Liu, and
  Lin]{xie2022optimization}
Xingyu Xie, Qiuhao Wang, Zenan Ling, Xia Li, Guangcan Liu, and Zhouchen Lin.
\newblock Optimization induced equilibrium networks: An explicit optimization
  perspective for understanding equilibrium models.
\newblock \emph{IEEE Transactions on Pattern Analysis and Machine
  Intelligence}, 2022.

\end{thebibliography}


\end{document}